\documentclass{article}
\usepackage[preprint]{neurips_2025}

\usepackage{enumitem}
\usepackage{natbib}
\usepackage{graphicx,wrapfig}
\usepackage[caption = false]{subfig}
\usepackage{subcaption}
\usepackage{dcolumn}
\usepackage[utf8]{inputenc}
\usepackage{blindtext}
\usepackage{xcolor}

\usepackage{tikz}
\usetikzlibrary{trees}
 \usepackage{soul}
 \usepackage{bm}
 \usepackage{mathtools,amsmath,amssymb,amsthm}
 \usepackage{float}
 \usepackage[caption = false]{subfig}
 \usepackage{graphicx}
\usepackage{multirow}
\usetikzlibrary{patterns,shapes.geometric, arrows.meta,positioning}

\theoremstyle{definition}

\title{Bias or Optimality? Disentangling Bayesian Inference and Learning Biases in Human Decision-Making}

\author{%
  Prakhar Godara \\
   Department of Psychology\\
  New York University\\
  New York, 10003 \\
  \texttt{prakhargodara@gmail.com} \\
}

\begin{document}
\maketitle

\begin{abstract}
Recent studies \citep{palminteri2017, Lefebvre2017, DENOUDEN2013, frank2007, chambon2020information, xia2021modeling} among others claim that human behavior in a two-armed Bernoulli bandit (TABB) task is described by \textit{positivity} and \textit{confirmation} biases, implying that "Humans do not integrate new information objectively." However, we find that even if the agent updates its belief via objective Bayesian inference, fitting the standard Q-learning model with asymmetric learning rates still recovers both biases. Bayesian inference cast as an effective Q-learning algorithm has symmetric, though decreasing, learning rates. We explain this by analyzing the stochastic dynamics of these learning systems using master equations: both confirmation bias and unbiased but decreasing learning rates yield the same behavioral signature—reduced action switching probabilities compared to constant, unbiased updates. Finally, we propose experimental protocols to disentangle true cognitive biases from artifacts of decreasing learning rates.
\end{abstract}


\section{Introduction}
{S}tudying human decision-making in idealized experimental settings has been a cornerstone of both behavioral economics \cite{Herrmann2008,Godara2022,Godara2023,Wang2024,andreoni2008} and cognitive science \cite{STEYVERS2009,zhang2013,wilson2014,BROWN2022,wu2022}. One particularly useful paradigm for such investigations is the $N$-armed Bernoulli bandit (called TABB when $N=2$) task \cite{Katehakis1987}, where participants must balance exploration and exploitation to maximize rewards. Each action (or arm) offers binary outcomes (zero or one) with unknown probabilities, and the subject’s goal is to maximize cumulative rewards over $T$ trials.

This task captures the explore-exploit dilemma faced by all decision-makers operating in uncertain environments. Its simplicity has made it a widely studied paradigm in behavioral studies, where researchers explore not only optimal decision strategies but also how real human behavior deviates from this ideal. Recently, the TABB task has also been used to study how people make inferences from stochastic feedback, and research has revealed the presence of cognitive biases in human learning—most notably, positivity and confirmation biases. These biases have repeatedly been observed in adult humans \citep{palminteri2017, Lefebvre2017, DENOUDEN2013,frank2007,chambon2020information}, adolescents \citep{xia2021modeling}, rodents \citep{OHTA2021218} and macaques \citep{farashahi2019flexible} to name a few. These findings are based on fitting a reinforcement learning framework, Q-learning, to subject behavior data.

Q-learning, which has been linked to dopaminergic activity in the brain \cite{dayan2005}, describes how agents update their expectations of reward from chosen and unchosen options based on the difference between expected and received outcomes. The model can be expressed as: \begin{equation}\label{eq:Q_learning} 
Q^v_{t+1} = Q^v_t + \begin{cases} \alpha^v_+ \cdot (r^v_t - Q^v_t), & \text{if } (r^v_t - Q^v_t) > 0 \\ \alpha^v_- \cdot (r^v_t - Q^v_t), & \text{if } (r^v_t - Q^v_t) < 0. \end{cases} \end{equation} where $v\in \{c,u\}$ denotes either the chosen arm ($c$) or the unchosen arm ($u$), and $r^v_t$ is the observed reward at time $t$ (corresponding to the $v$ arm). Here we have written the Q-learning in a general manner, accounting for tasks with (called "Experiment-2" in \cite{palminteri2017}) and without ("Experiment-1") counterfactual information\footnote{Usually the agents are only informed of the rewards of the chosen arm i.e. no counterfactual information.}. In particular for tasks without counterfactual information we have $\alpha^u_\pm = 0$.

The parameters $\alpha^v_+$ (for positive prediction errors) and $\alpha^v_-$ (for negative prediction errors)  are learning rates, dictating how quickly the agent adjusts its expectations based on the outcome. Human biases are formalized through these learning rates: \begin{enumerate}
    \item Positivity bias: $\alpha^c_+ > \alpha^c_-$, i.e. the agent, for the chosen arm, favors positive updates more than negative ones.
    \item Confirmation bias: $\alpha^c_+ > \alpha^c_-$ and $\alpha^u_- > \alpha^u_+$, i.e. the agent favors positive updates for the chosen arm, but negative updates for the unchosen arm.
\end{enumerate} 

However, Q-learning is not the only framework used to model human behavior. Another influential approach is Bayesian inference, in which agents update their beliefs in a statistically optimal manner based on the likelihood of new evidence. Because Bayesian models are often considered the benchmark of rational inference \cite{zellner1988}, they have become central to many cognitive science models of decision-making \cite{George2009,Griffiths2008}. While Bayesian inference operates differently from Q-learning, one might naturally expect that if we fit a Q-learning model to data generated by a Bayesian agent, the inferred learning rates would be unbiased—after all, a rational learner should not exhibit systematic distortions. In other words, if our methodology for detecting biases is sound, it should classify Bayesian inference as an unbiased process. Yet, as we will show, this is not the case.

When fitting the Q-learning model to data generated by an agent using Bayesian inference, we still observe the same biases—positivity and confirmation—commonly found in human data. This result raises an important question: are these biases inherent to human cognition, or do they emerge as an artifact of the Q-learning framework used to model behavior?\footnote{A similar study \citep{KATAHIRA201831} studies how a choice autocorrelation can generate pseudo-positivity/confirmation bias. Here we present yet another mechanism that generates such statistical artefacts.}

In this paper, we demonstrate that the latter is true. In particular we will focus on TABB tasks with counterfactual information as both positivity and confirmation biases are observed in that setup. First, we will reformulate Bayesian inference in terms of Q-learning to establish an equivalent framework. We show that while Bayesian inference leads to unbiased learning, the learning rates decrease over time, which can lead to the appearance of biases when fitting constant learning rate models. Next, we will use the master equation approach \cite{gardiner1985} to study the stochastic dynamics of these learning systems and provide an explanation for the emergent biases. Finally, we will compare the predictive power of Bayesian models versus Q-learning models in fitting human behavioral data from \cite{palminteri2017}, and propose novel experiments that could help disentangle true cognitive biases from optimal learning mechanisms.

\section*{Bayesian inference models}
\subsection*{Bayes-greedy and Bayes-optimal behavior in TABB}
We begin by describing the Bayes-optimal behavior for the TABB task of a finite horizon $T$ and no counterfactual information. As the rewards corresponding to a single arm (or action) $a=i\in \{1,2\}$ in TABB are Bernoulli random variables   parameterized by $p_i$, the belief of the agent about the reward rates can be viewed as a probability distribution over possible values of $p_i$.  Assuming that the agent performs Bayesian inference and starts with a uniform prior distribution over $p_i\in [0,1]$ for $i\in \{1,2\}$, the belief probability distributions can be represented by the two-parameter beta distribution. At any given time $t$, the belief\footnote{Here we tacitly identify the belief distribution with the parameters of the belief distribution.} can then be written as \begin{equation}
    b_t = (\alpha_1, \beta_1, \alpha_2, \beta_2) \in \mathbb{N}_0^4 : \alpha_1+ \beta_1+ \alpha_2+ \beta_2 = t ,
\end{equation} where $\alpha_i$ is the number of (thus far) obtained successes corresponding to pulling the arm $a=i$, and $\beta_i$ the failures. Here $(\alpha_i,\beta_i)$ are the parameters of the beta-distribution corresponding to arm $a=i$. We therefore see that in the specific case of TABB, Bayes inference can be viewed as a deterministic update rule \begin{equation}\label{eq:update_bt}
    b_{t+1} = b_t + y (a_t,r_t),
\end{equation}where $r_t$ is the reward obtained at time $t$ for the chosen arm, and $y$ is given by\[   y(a,r) = \begin{cases}
   (1,0,0,0), &\text{if } a=1,r=1, \\
   (0,1,0,0), &\text{if } a=1,r=0, \\
   (0,0,1,0), &\text{if } a=2,r=1, \\
   (0,0,0,1), &\text{if } a=2,r=0.
\end{cases} \] Further, the mean of the beta distribution with parameters $(\alpha_i,\beta_i)$ is given by $p_i^s = \frac{\alpha_i+1}{\alpha_i+\beta_i+2}$. This represents the "average/effective belief" of the agent about the reward of an arm, i.e. the agent effectively believes the reward to be distributed according to a Bernoulli distribution with parameter $p_i^s$ (here $\cdot^s$ is used to demonstrate that this is the subjective $p_i$ value), i.e., the distribution of rewards under the average belief $b_t(r|a=i) = \mathrm{B}(p_i^s)$.

So far we described how the agent updates its beliefs. We now proceed to describe the decision policy $a_t = \pi(b_t)$. The (Bayes-)optimal decision policy $\pi^*$ is a solution of the Bellman equation as follows\footnote{This is also known in the literature as a (special case of) Bayes Adaptive Markov Decision Process (BAMDP) \cite{duff2002}.}. \begin{equation}\label{eq:Bayes-optimal}
    V^*(b_{t}) =  \max_a  \sum_r \Big[b_t(r|a)\big(r +\delta (y-y(a,r)) V(b_{t}+y)\big)   \Big],
\end{equation} where $\delta$ represents the Dirac-delta function, and $\pi^*(b_t) = \arg\max_a V(b_t)$. Notice that in this strategy, we maximize a sum of immediate and future rewards. For tasks with counterfactual information, where the agents are also presented with the evidence of the reward of the unchosen arm,  the subsequent belief states become independent of the current action, I.e. $y$ becomes independent of $a$, further implying that the second summand just acts as a constant (w.r.t. $a$) and can therefore be removed from Eq. \ref{eq:Bayes-optimal}. Therefore, in presence of counterfactual evidence, the Bellman equation becomes \begin{equation}\label{eq:Bayes-greedy}
    V^{*}(b_t) = \max_a \sum_rb_t(r|a)r.
\end{equation}

\subsection*{Mapping Bayesian inference to $Q$-learning}
In order for Bayesian inference and $Q$-learning to be comparable, we need to be able to treat the two formulations in an equal setting, i.e. draw a correspondence between $Q$-learning (Eq. \ref{eq:Q_learning}) and Bayesian inference (Eqs. \ref{eq:Bayes-optimal},\ref{eq:Bayes-greedy}). We notice that the action chosen in the $Q$-learning paradigm is a soft-max of the two $Q$-values, suggesting that the $Q$ values act as a prediction for the expected reward of the two arms. Additionally, as the $Q$-values always stay in the interval $[0,1]$, it seems appropriate to equate them to the estimates of immediate expected rewards $p_i$ of the two arms. 

Using the update rule for $b_t$ (Eq. \ref{eq:update_bt}), we can identify an similar update rule for $p^s_{i,t}$ as\[
p^s_{i,t+1} = 
\begin{cases} 
\frac{\alpha_{i,t} + 2}{\alpha_{i,t} + \beta_{i,t} + 3}, & \text{if } a_t = i, r_t = 1, \\ 
\frac{\alpha_{i,t} + 1}{\alpha_{i,t} + \beta_{i,t} + 3}, & \text{if } a_t = i, r_t = 0, \\ 
\frac{\alpha_{i,t} + 1}{\alpha_{i,t} + \beta_{i,t} + 2}, & \text{if } a_t \neq i.
\end{cases}
\] Given that we map $p^s_{i,t}$ to $Q^v_t$, let us re-write the above evolution equation in a manner similar to Eq. \ref{eq:Q_learning}. We then write \begin{equation}
    p^s_{v,t+1} = p^s_{v,t} + \begin{cases}
        \alpha^v_+ \cdot (r^v_t - p^s_{v,t}), & \text{if} (r^v_t = 1), \\
         \alpha^v_- \cdot (r^v_t - p^s_{v,t}), & \text{if} (r^v_t = 0). 
    \end{cases}
\end{equation} This implies that $\alpha_+ = \frac{p^s_{v,t+1}-p^s_{v,t}}{1-p^s_{v,t}}$ where $p^s_{v,t+1}$ is given by the update rule with $r_t = 1$ and $\alpha_- = \frac{p^s_{v,t+1}-p^s_{v,t}}{-p^s_{v,t}}$ with $p^s_{v,t}$ is given by the update rule with $r_t = 0$. Upon making this substitution we find that\begin{equation}\label{eq:effective_learning_Bayes}
    \alpha_\pm = \frac{1}{\alpha_{v,t} + \beta_{v,t} + 3} = \frac{1}{t+3}.
\end{equation} 

This result is of fundamental importance. We have established that Bayesian inference can be precisely mapped onto a Q-learning formulation with no inherent bias (i.e., $\alpha^v_+ = \alpha^v_-$)\footnote{This result holds specifically for Bernoulli bandits; although an analogous derivation applies to Gaussian bandits under a Gaussian prior. More general relationships between Bayesian inference and Q-learning in broader settings remain an open question.}. Given this, the findings from the model-fitting procedure in Fig. \ref{fig:data_present}(a) present a significant conceptual challenge. The behavioral data are generated from an unbiased algorithm—Bayesian inference with optimal decision-making—yet, when fitting the Q-learning model (Eq. \ref{eq:Q_learning}) to this data, the inferred learning rates exhibit asymmetry, suggestive of bias.

This discrepancy suggests that the source of the observed biases does not lie in the underlying decision process itself but rather in the structure of the model used to analyze it. Specifically, while Bayesian inference follows a learning rate that decreases monotonically as $\sim \frac{1}{t}$ (where $t$ denotes the number of pulls), the standard Q-learning framework assumes a constant learning rate. As a result, when fitting Q-learning to Bayesian-generated data, the failure to account for this temporal dependence leads to the systematic misestimation of learning rates, introducing spurious asymmetries. In the following section, we provide a formal analysis of this phenomenon and its implications for modeling human decision-making.

\section{Explaining the observation}

\begin{figure}
\centering
\subfloat[]{\includegraphics[width = 1.7in]{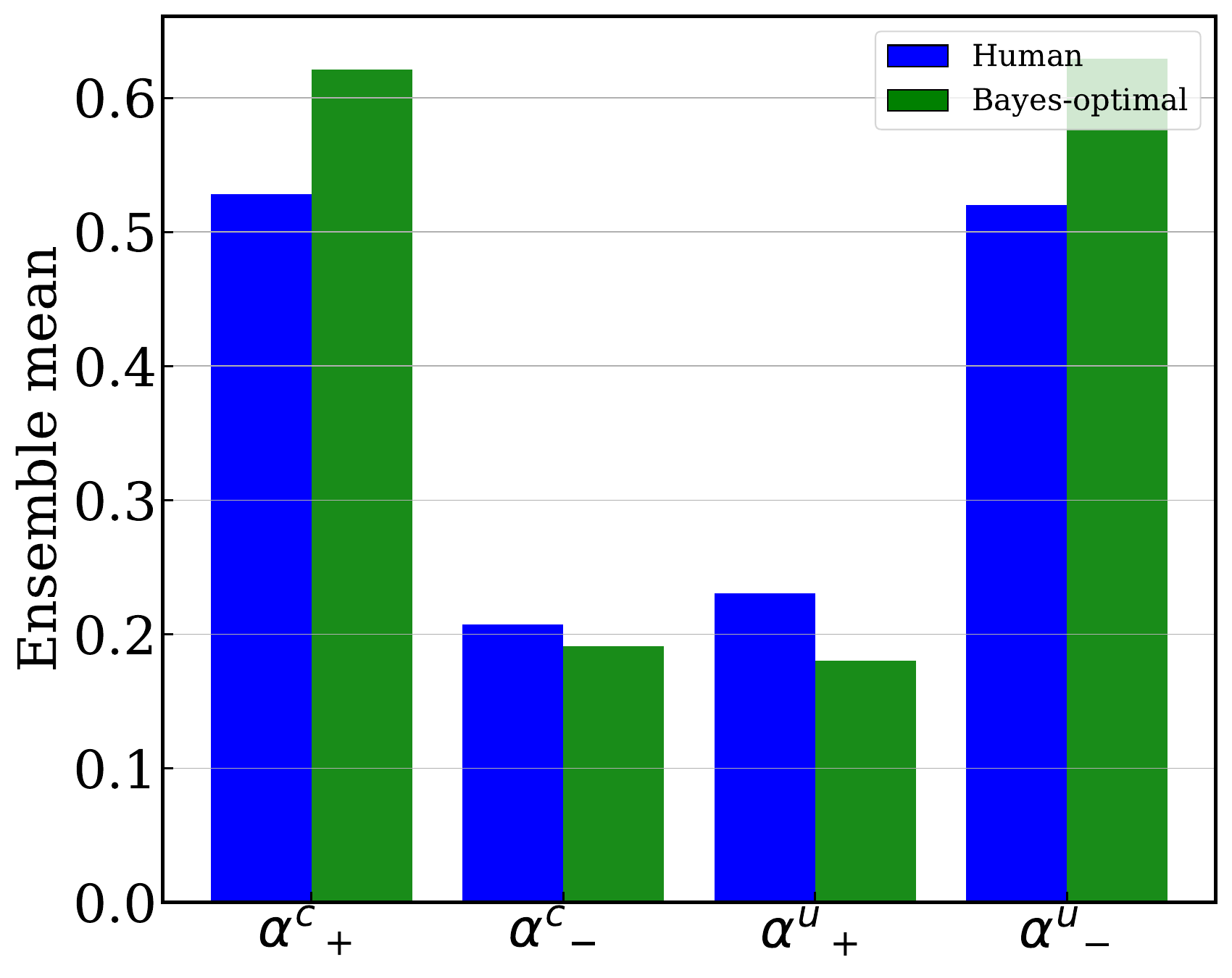}} 
\subfloat[]{\includegraphics[width = 1.7in]{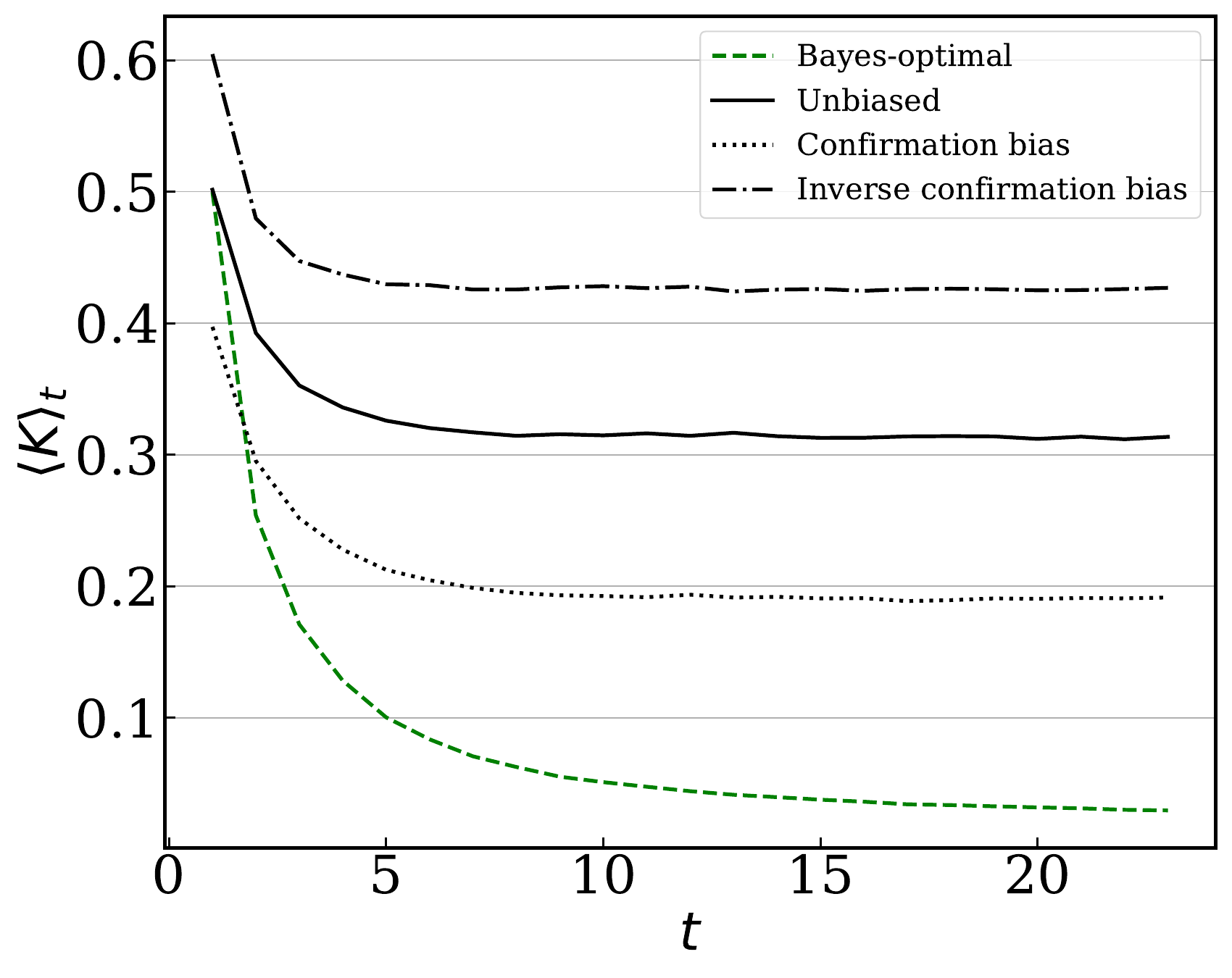}}
\caption{(a) Ensemble averaged best fit learning rates when fit to human data \cite{palminteri2017} (blue) and Bayes-optimal agents (green). (b) Ensemble averaged action switching rates $\langle K \rangle_t$ as a function of time for $Q$-learning algorithms (black), humans (blue) and Bayes-optimal policy (green).} 
\label{fig:data_present}
\end{figure}

With the correspondence between $Q$-learning and Bayes-greedy agents in place, we can now begin to compare the two $Q$-learning algorithms. More specifically we ask the question - when fitting stationary learning rates (from Eq. \ref{eq:Q_learning}) to data generated by Eq. \ref{eq:effective_learning_Bayes}, why do we observe a confirmation/positivity bias? Further, as both positivity and confirmation biases are observed in TABB tasks with counterfactual information, we will perform our analysis only for that setting. A similar analysis can be performed for the case without counterfactual evidence.

We will argue that the answer lies in the dynamics of action switching probabilities implied by the two algorithms (see Fig. \ref{fig:data_present} (b)). More specifically, we will show that, when starting from constant and unbiased learning rates, there are two distinct ways of reducing the action switching rates - either by introducing a (positivity/confirmation) bias and keeping the learning rates constant or by maintaining unbiasedness in the learning rates but decreasing them monotonically with time\footnote{Not all temporal protocols lead to a decrease in action switching rates.}. Therefore, we begin by first exploring the effects of biases on action switching rates, then followed by temporally varying learning rates.  Finally, as the action switching rates also depend on the reward statistics of the environment, we consider symmetric environments i.e. $(p_1,p_2) = (p,p)$. This not only simplifies the analytical calculations but also allows us the study the exclusive effects of biases on action switching rates.

\subsection{Action switching under constant learning rates} \label{sec:biased_explanation}

We now start with the $Q$-learning algorithm from Eq. \ref{eq:Q_learning}, where the agent is described by the parameters $(\{\alpha^v_\pm\},\beta)$. Eq. \ref{eq:Q_learning} is a stochastic difference equation (SDE) governing the time evolution $\bm Q = (Q_1,Q_2)$. Obtaining relevant quantities of interest would require us to perform multiple simulation runs (for averaging) of the Q-learning algorithm. This becomes  particularly problematic if we are to explore a high dimensional parameter space (five dimensional system from Eq. \ref{eq:Q_learning}). Alternatively, we could directly consider the time evolution of the probability distribution $P(\bm Q)$ over the Q-values, via what is called a \textit{master equation} \cite{gardiner1985}. This allows us to consider the time evolution of a deterministic system, and also allows further analytical treatment.

We denote the distribution at time $t$ by $P_t(\cdot)$ and therefore write the master equation as \begin{equation}\label{eq:master_equation}
    P_{t+1}(\bm Q) = \int P_t(\bm Q') \mathcal{T}(\bm Q' \to \bm Q; \{\alpha^v_\pm\}) d\bm Q',
\end{equation} where $\mathcal{T}$ denotes the transition kernel of the stochastic process. In our case, $\mathcal{T}(\bm Q' \to \bm Q; \{\alpha^v_\pm\})$ is given by \begin{equation}\label{eq:transition_kernel}
    \begin{split}
        \mathcal{T}(\bm Q' \to \bm Q; \{\alpha^v_\pm\}) = & \sum_i \pi_i(\bm Q')  \mathcal{T}_{-i}(Q_{-i},Q'_{-i};\alpha^u_\pm)    \mathcal{T}_i(Q_i,Q'_i;\alpha^c_\pm),
    \end{split}
\end{equation} where $\pi_i$ is the probability of taking action $i$, $\mathcal{T}_{-i} \equiv p\delta (Q_{-i} - Q'_{-i} - \alpha^u_+(1-Q'_{-i})) + (1-p)\delta (Q_{-i} - Q'_{-i} + \alpha^u_-Q'_{-i})$ and $\mathcal{T}_i \equiv p\delta (Q_i - Q'_i - \alpha^c_+(1-Q'_i)) + (1-p)\delta (Q_i - Q'_i + \alpha^c_-Q'_i)$. Here the product $\mathcal{T}_{-i}\mathcal{T}_{i}$ reflects the assumption of independent rewards for both the arms and the specific forms reflect the update rules given by Eq. \ref{eq:Q_learning} with the reward probabilities $(p_1,p_2) = (p,p)$ as aforementioned.

We are interested in the total probability of switching the action under a 1-step transition from the state $\bm Q'$. In order to obtain that we first write the action switching probability under a specific transition  $\bm Q' \to \bm Q$, which we represent by $K(\bm Q'\to \bm Q;\{\alpha^v_\pm\})$ and is given by $K(\bm Q'\to \bm Q;\{\alpha^v_\pm\}) =   \sum_i \pi_i(\bm Q')  (1-\pi_i(\bm Q))  \mathcal{T}_{-i}(Q_{-i},Q'_{-i};\alpha^u_\pm) \mathcal{T}_i(Q_i,Q'_i;\alpha^c_\pm)$. The total probability of action switching is then obtained by marginalising over the future states $\bm Q$ and is given by \begin{equation}\label{eq:act_switch_prob}
    \begin{split}
        K(\bm Q';\{\alpha^v_\pm\}) = &   \sum_i \pi_i(\bm Q') \int (1-\pi_i(\bm Q))  \mathcal{T}_{-i}(Q_{-i},Q'_{-i}; \alpha^u_\pm  )  \mathcal{T}_i(Q_i,Q'_i;\alpha^c_\pm) d\bm Q.
    \end{split}
\end{equation} The ensemble-average total probability of action switching at time $t$ is then given by \begin{equation}\label{eq:avg_act_switch_prob}
    \langle K \rangle_t = \int K(\bm Q;\{\alpha^v_\pm\}) P_t(\bm Q) d\bm Q.
\end{equation}

We can obtain the time-evolution of the ensemble-average of a quantity $f(\bm Q)$ as \begin{equation} \label{eq:average_evolution_f}
    \begin{split}
        \langle f \rangle_{t+1} &= \int f(\bm Q) P_{t+1}(\bm Q) d\bm Q  \\ & = \int P_t(\bm Q') \int f(\bm Q) \mathcal{T}(\bm Q' \to \bm Q ; \{\alpha_\pm^v\}) d\bm Q d \bm Q' = \langle I \rangle_t,
    \end{split}
\end{equation} where $I = \int f(\bm Q)\mathcal{T}(\bm Q' \to \bm Q ; \{\alpha_\pm^v\})d\bm Q$. Making the substitution for $\mathcal{T}$ we get \begin{equation}\begin{split}
    \langle f\rangle_{t+1} =&  \langle p^2f(Q_1^{+,u},Q_2^{+,c}) +  p(1-p)f(Q_1^{-,u},Q_2^{+,c})  \\ & + p(1-p)f(Q_1^{+,u},Q_2^{-,c}) +  (1-p)^2f(Q_1^{-,u},Q_2^{-,c})\rangle_t \\&
+   \langle\pi [ p^2f(Q_1^{+,c},Q_2^{+,u}) + p(1-p)f(Q_1^{-,c},Q_2^{+,u})   \\ & + p(1-p)f(Q_1^{+,c},Q_2^{-,u}) + (1-p)^2f(Q_1^{-,c},Q_2^{-,u}) \\ & -  p^2f(Q_1^{+,u},Q_2^{+,c}) -  p(1-p)f(Q_1^{-,u},Q_2^{+,c})  \\ & - p(1-p)f(Q_1^{+,u},Q_2^{-,c}) - (1-p)^2f(Q_1^{-,u},Q_2^{-,c}) ] \rangle_t,   
\end{split}
\end{equation} and here we use the shorthand $Q_i^{+,v} = Q_i + \alpha^v_+ (1-Q_i)$ and $Q_i^{-,v} = Q_i - \alpha^v_- Q_i$. As we are dealing with symmetric environments we have $\langle f(Q_1,Q_2) \rangle = \langle f(Q_2,Q_1)\rangle$. Therefore, by making use of Eq. \ref{eq:average_evolution_f} we can obtain the time evolution equations for the first two moments of $P(Q_1,Q_2)$ as follows. For the mean $\langle Q_1 \rangle$ we have \begin{equation} \label{eq:q1_exact}
   \langle Q_1 \rangle_{t+1} = B^{(1)}_1\langle Q_1 \rangle_t  
+ B^{(2)} \langle \pi \rangle_t  + B^{(2)}_1 \langle\pi Q_1 \rangle_t + B^{(1)}.
\end{equation} Here, $B^{(1)}_1 = p^2(1-\alpha_+^u) + (1-p)^2 (1-\alpha_-^u) + p(1-p)(2-\alpha_+^u-\alpha_-^u)$, $B^{(2)}_1 = p^2 (\alpha_+^u-\alpha_+^c) + (1-p)^2 (\alpha_-^u-\alpha_-^c) + p(1-p) (\alpha_+^u + \alpha_-^u-\alpha_-^c-\alpha_+^c)$, $B^{(2)} = p(\alpha_+^c-\alpha_+^u)$ and $B^{(1)} = p\alpha^u_+$. Similarly $\langle Q_1Q_2 \rangle$ evolves according to \begin{equation}\begin{split}
        \langle Q_1Q_2\rangle_{t+1} =& \bigg[C^{(1)}_{12}\langle Q_1Q_2\rangle_{t} + C^{(1)}_{1}\langle Q_1\rangle_{t} + C^{(1)}_{2}\langle Q_2\rangle_{t} + C^{(1)}\bigg] + 
        \\ & \bigg[\big(C^{(2)}_{12}-C^{(1)}_{12}\big)\langle \pi Q_1Q_2\rangle_{t} + \big(C^{(2)}_{1}-C^{(1)}_{1}\big)\langle \pi Q_1\rangle_{t} + \\& \big(C^{(2)}_{2}-C^{(1)}_{2}\big)\langle \pi Q_2\rangle_{t} + \big(C^{(2)}-C^{(1)}\big)\langle \pi \rangle_t \bigg].
\end{split}
\end{equation} Here $C^{(1)}_{12} = p^2(1-\alpha^u_+)(1-\alpha^c_+) + p(1-p)\big[ (1-\alpha^u_-)(1-\alpha^c_+)+ (1-\alpha^u_+)(1-\alpha^c_-)\big] + (1-p)^2(1-\alpha^u_-)(1-\alpha^c_-) = C^{(2)}_{12}$, $C^{(1)}_{1} =\alpha_+^c (1 - \alpha^u_+)p^2 + \alpha_+^c (1 - \alpha^u_-)(1-p)p = C^{(2)}_2$, $C^{(1)}_{2} =\alpha_+^u (1 - \alpha^c_+)p^2 + \alpha_+^u (1 - \alpha^c_-)(1-p)p = C^{(2)}_1$ and $C^{(1)} = p^2 \alpha^c_+ \alpha^u_+ = C^{(2)}$. Using the relationships between these constants and also making use of the symmetry relations - $\langle \pi Q_1 \rangle = \langle (1-\pi) Q_2 \rangle$ and $\langle Q_1 \rangle = \langle Q_2 \rangle$ we can simplify the above expression to \begin{equation} \label{eq:q1q2_exact}
    \begin{split}
         \langle Q_1Q_2\rangle_{t+1} = & C^{(1)}_{12}\langle Q_1Q_2\rangle_{t} + \big(C^{(1)}_{1}+ C^{(1)}_{2}\big)\langle Q_1\rangle_{t}  \\ & + C^{(1)} +  \big(C^{(1)}_2 - C^{(1)}_1\big)\bigg[ 2\langle \pi Q_1\rangle_t - \langle Q_1\rangle_t \bigg].
    \end{split}
\end{equation} Finally $\langle Q^2_1\rangle$ evolves according to \begin{equation}\label{eq:q1q1_exact}
    \begin{split}
        \langle Q^2_1\rangle_{t+1} & =\bigg[ D^{(1)}_{11} \langle Q^2_1\rangle_t + D^{(1)}_{1} \langle Q_1\rangle_t + D^{(1)}  \bigg]   \\& + \bigg[ (D^{(2)}_{11} - D^{(1)}_{11})\langle \pi Q_1^2\rangle_t  + (D^{(2)}_1 - D^{(1)}_1)\langle \pi Q_1\rangle_t  \\ & + (D^{(2)} - D^{(1)})\langle \pi \rangle_t \bigg],
    \end{split}
\end{equation} where $D^{(1)}_{11} = p(1-\alpha^u_+)^2 + (1-p)(1-\alpha^u_-)^2$, $D^{(1)}_{1} = 2p\alpha^u_+(1-\alpha^u_+)$, $D^{(1)} = p(\alpha^u_+)^2$ and the corresponding $D^{(2)}$ terms are the same, except that $\cdot^u$ is replaced with $\cdot^c$.

The above three deterministic difference equations in Eqs. \ref{eq:q1_exact}, \ref{eq:q1q2_exact} and \ref{eq:q1q1_exact} govern the time evolution of the first two moments of the joint probability distribution over the Q-values. However, they require us to know the integrals for $\langle \pi \rangle_t,\langle\pi Q_1 \rangle_t$ and $\langle \pi Q^2_1 \rangle_t$. Generally speaking, evaluating these integrals in closed form is only possible in very simplified scenarios and sadly Eq. \ref{eq:master_equation} doesn't fall in that category. One can however, recall the standard textbook knowledge (\cite{benaroya2005}) that the mean of some (non-linear) transformation (in our case $f$) of a random variable ($\bm Q$ in our case) can be approximated by transformations of the moments of the random variable. This should simplify our lives and allow us to approximate the integrals above and consequently provide us with (approximate) deterministic dynamics for $\langle f \rangle_t$ which would be amenable to deeper analytical treatment in comparison to stochastic difference equations as in Eq. \ref{eq:Q_learning}. The approximation can be derived as follows. \begin{equation}
   \int f(\bm Q) P_t(\bm Q) d\bm Q = \int f(\bm \mu_Q + (\bm Q - \bm \mu_Q)) P_t(\bm Q) d\bm Q. 
\end{equation} If $f$ is a smooth function, we can Taylor expand it around $\bm \mu_Q = (\langle Q_1 \rangle, \langle Q_2 \rangle)$ upto second order gives, \begin{equation}
 \begin{split}
     f(\bm \mu_Q + (\bm Q - \bm \mu_Q)) & \approx f(\bm \mu_Q) + \bm \nabla f(\bm \mu_Q) \cdot (\bm Q - \bm \mu_Q) \\ & + \frac{1}{2!} (\bm Q - \bm \mu_Q)^T H_f (\bm Q - \bm \mu_Q) \\ & + \mathcal{O}((\bm Q- \bm \mu_Q)^3),
 \end{split}   
\end{equation} where $H_f$ is the Hessian matrix of $f$. Finally, performing the integral over this approximation of $f$ then  allows us to write
\begin{equation} \label{eq:moment_approximation}
 \langle f \rangle_t \approx f(\bm \mu_Q) + \frac{1}{2} H_{f,ij} \langle (Q_i - \mu_i)(Q_j - \mu_j) \rangle_t,   
\end{equation} where we've used the Einstein summation convention. Eq. \ref{eq:moment_approximation} finally allows us to approximate the mean of a non-linear transformation $f$ in terms of mean and covariance matrices of $\bm Q$. We can make use of this approximation to now write down the integrals $\langle \pi \rangle_t,\langle\pi Q_1 \rangle_t$ and $\langle \pi Q^2_1 \rangle_t$ in terms of $\langle Q_1 \rangle_t$, $\langle Q_1Q_2 \rangle_t$ and $\langle Q^2_1 \rangle_t$, the evolution equations of which are have already been obtained  in Eqs. \ref{eq:q1_exact}, \ref{eq:q1q2_exact} and \ref{eq:q1q1_exact} respectively.

We begin by evaluating $\langle \pi \rangle_t$. Making use of the functional form of $\pi (\bm Q) = \frac{1}{1 + e^{-\beta(Q_1-Q_2)}}$ we can obtain its Hessian matrix to be \begin{equation}
    H_\pi =
\beta^2 \pi (1 - \pi) (1 - 2\pi)
\begin{pmatrix}
1 & -1 \\
-1 & 1
\end{pmatrix}.
\end{equation} Recalling the symmetry in our system we have $\langle Q_1 \rangle_t - \langle Q_2 \rangle_t$, implying that $\pi(\bm \mu_Q) = \frac{1}{2}$ and therefore $H_\pi = \bm 0$, making the second term vanish. This finally implies $\langle \pi \rangle_t = \frac{1}{2}$.

For $\langle \pi Q_1 \rangle_t$ we need to evaluate the Hessian matrix $H_{\pi Q_1}$, which evaluates to \begin{equation}
    H_{\pi Q_1} = \beta\pi (1 - \pi)
\begin{pmatrix}
2  + \beta Q_1 (1 - 2\pi) & -1 - \beta Q_1 (1 - 2\pi) \\
-1  - \beta Q_1 (1 - 2\pi) & \beta Q_1 (1 - 2\pi)
\end{pmatrix},
\end{equation} which under the symmetry condition evaluates to \begin{equation}
    H_{\pi Q_1} = \frac{\beta}{4}
\begin{pmatrix}
2   & -1 \\
-1   & 0
\end{pmatrix}.
\end{equation} Making use of this we obtain the approximation for  $\langle \pi Q_1 \rangle_t$ to be \begin{equation}
    \langle \pi Q_1 \rangle_t = \frac{\langle Q_1\rangle_t}{2} + \frac{\beta}{4} (\langle Q_1^2\rangle_t - \langle Q_1Q_2\rangle_t)
\end{equation}

Finally, $H_{\pi Q^2_1}$ evaluates to \begin{equation}
H_{\pi Q^2_1} = \pi 
\begin{bmatrix}
\shortstack[l] 
{$2 + 2\beta Q_1 (1 - \pi) + \beta (1-\pi)$\\[0pt]
$\quad (2Q_1 +\beta Q^2_1(1-2\pi))$}
&
\shortstack[l]{
$- \beta (1 - \pi) (2Q_1 +$\\[0pt]
$\quad \beta Q^2_1 (1-2\pi))$
}\\[10pt]
\shortstack[l]{
$- \beta (1 - \pi) (2Q_1 +$\\[0pt]
$\quad \beta Q^2_1 (1-2\pi))$
}
&
\beta Q^2_1 (1-\pi)(1 - 2\pi)
\end{bmatrix}
\end{equation} which under the symmetry condition evaluates to \begin{equation}
    H_{\pi Q^2_1} = 
\begin{pmatrix}
1  + \beta Q_1  & -\frac{\beta}{2} Q_1 \\
 -\frac{\beta}{2} Q_1 & 0
\end{pmatrix}.
\end{equation} This gives us approximation \begin{equation}
    \langle \pi Q^2_1 \rangle_t = \frac{\langle Q^2_1 \rangle_t}{2} + \frac{\beta}{2} \langle Q_1\rangle_t (\langle Q_1^2\rangle_t - \langle Q_1Q_2\rangle_t).
\end{equation}

We can now substitute these approximations into Eqs. \ref{eq:q1_exact}, \ref{eq:q1q2_exact} and \ref{eq:q1q1_exact} to obtain the closed-form (approximate) dynamics for the first two moments to be \begin{subequations}\label{eq:moments_evolution_final}
  \begin{equation}
    \label{eq-a}
       \langle Q_1 \rangle_{t+1} = \big(B^{(1)}_1 + \frac{B^{(2)}_1}{2}\big)\langle Q_1 \rangle_t  
 + \frac{B^{(2)}_1\beta}{4} \Delta_t + B^{(1)} + \frac{B^{(2)}}{2} .
  \end{equation}
  \begin{equation}
    \label{eq-b}
   \begin{split}
       \langle Q_1Q_2\rangle_{t+1} & = C^{(1)}_{12}\langle Q_1Q_2\rangle_{t} + \big(C^{(1)}_{1}+ C^{(1)}_{2}\big)\langle Q_1\rangle_{t} \\ & + C^{(1)} +  \big(C^{(1)}_2 - C^{(1)}_1\big)\frac{\beta}{2}\Delta_t,
   \end{split}
  \end{equation}
\begin{equation}
    \label{eq-c}
     \begin{split}
        \langle Q^2_1\rangle_{t+1} & = \bigg[ \frac{(D^{(2)}_{11} + D^{(1)}_{11})}{2} \langle Q^2_1\rangle_t + \frac{(D^{(2)}_1 + D^{(1)}_1)}{2} \langle Q_1\rangle_t + D^{(1)}  \bigg]   \\& +  \bigg[ (D^{(2)}_{11} - D^{(1)}_{11})\langle Q_1\rangle_t\frac{\beta}{2}\Delta_t \\ & + (D^{(2)}_1 - D^{(1)}_1)\frac{\beta}{4}\Delta_t + \frac{(D^{(2)} - D^{(1)})}{2} \bigg],
    \end{split}
  \end{equation}
\end{subequations} Here we write $\langle \frac{(Q_1-Q_2)^2}{2}\rangle_t = \langle Q_1^2\rangle_t - \langle Q_1Q_2\rangle_t = \Delta_t$.

With the approximate solution of the master equation in place, we are now prepared to comment on the time evolution of $\langle K \rangle_t$. Instead of going from the route of Eq. \ref{eq:moment_approximation} which would be cumbersome, we start by first taking a closer look at the form of $K$ from Eq. \ref{eq:act_switch_prob}. Some simple algebra reveals that $K$ is monotonically decreasing in $\sqrt{2\Delta} = |Q_1-Q_2|,\alpha_+^c,\alpha_-^u$ and monotonically increasing in $\alpha_-^c,\alpha_+^u$. Intuitively, as the $Q$ values go far apart, the probability of switching the action in just one time-step is lower. Similarly, higher values of  $\alpha_+^c,\alpha_-^u$ drive the $Q$ values of the chosen and the unchosen options further apart (as the chosen option is more likely going to be the bigger of the $Q$ values), thereby decreasing the action switching probability. And an opposite effect is to be expected for  $\alpha_-^c,\alpha_+^u$. As we wish to look for the effect of constant learning rates on the temporal evolution of  $\langle K \rangle_t$, the temporal variation is only going to come via the (temporal-)variation in $\Delta_t$. We can already make use of Eqs. \ref{eq:moments_evolution_final} to find the time evolution of $\Delta_t$ and more specifically look at the steady state value $\Delta^*$ and note how it varies with confirmation bias. 

\begin{figure}
\centering
\subfloat[]{\includegraphics[width = 1.7in]{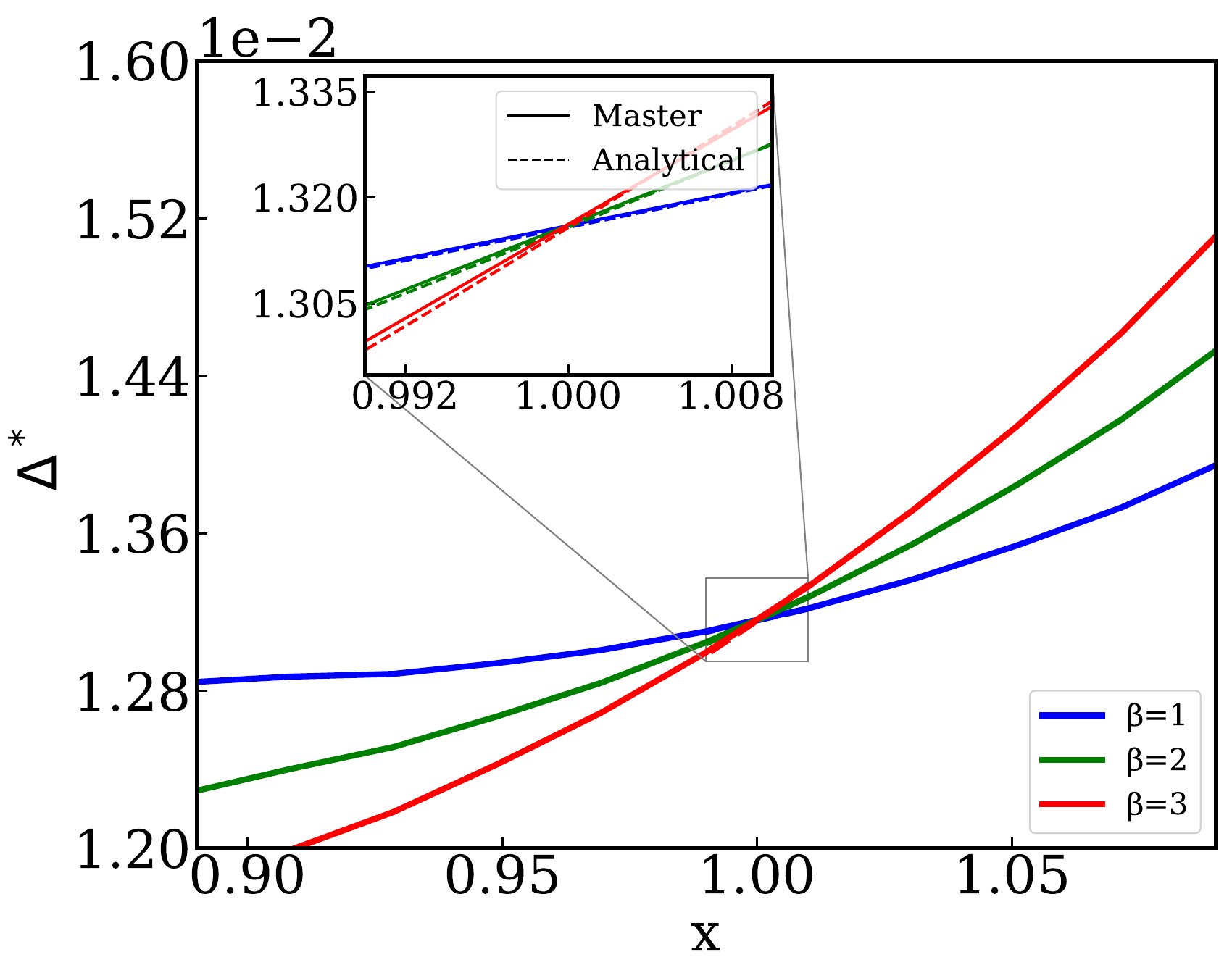}} 
\subfloat[]{\includegraphics[width = 1.7in]{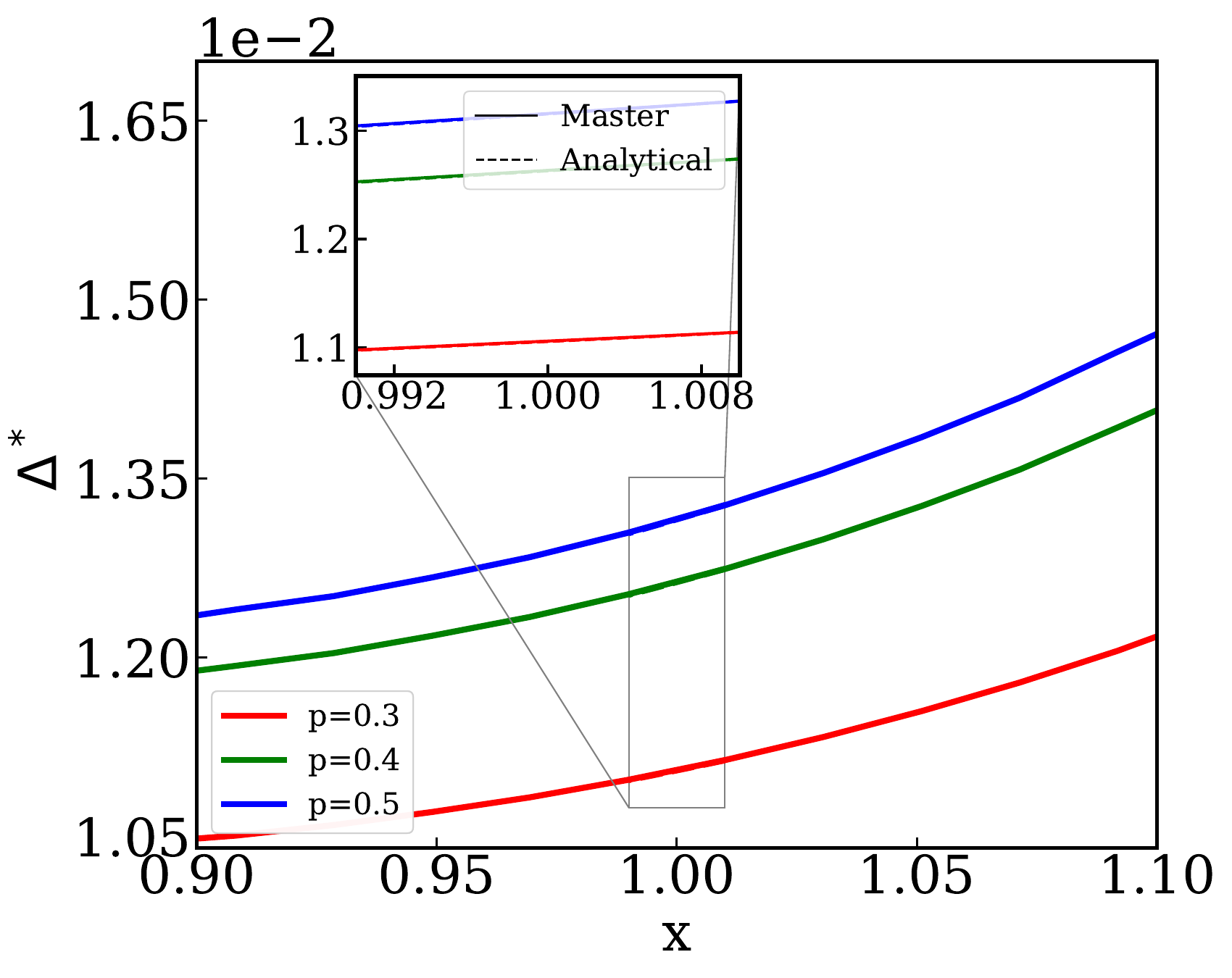}}
\subfloat[]{\includegraphics[width = 1.7in]{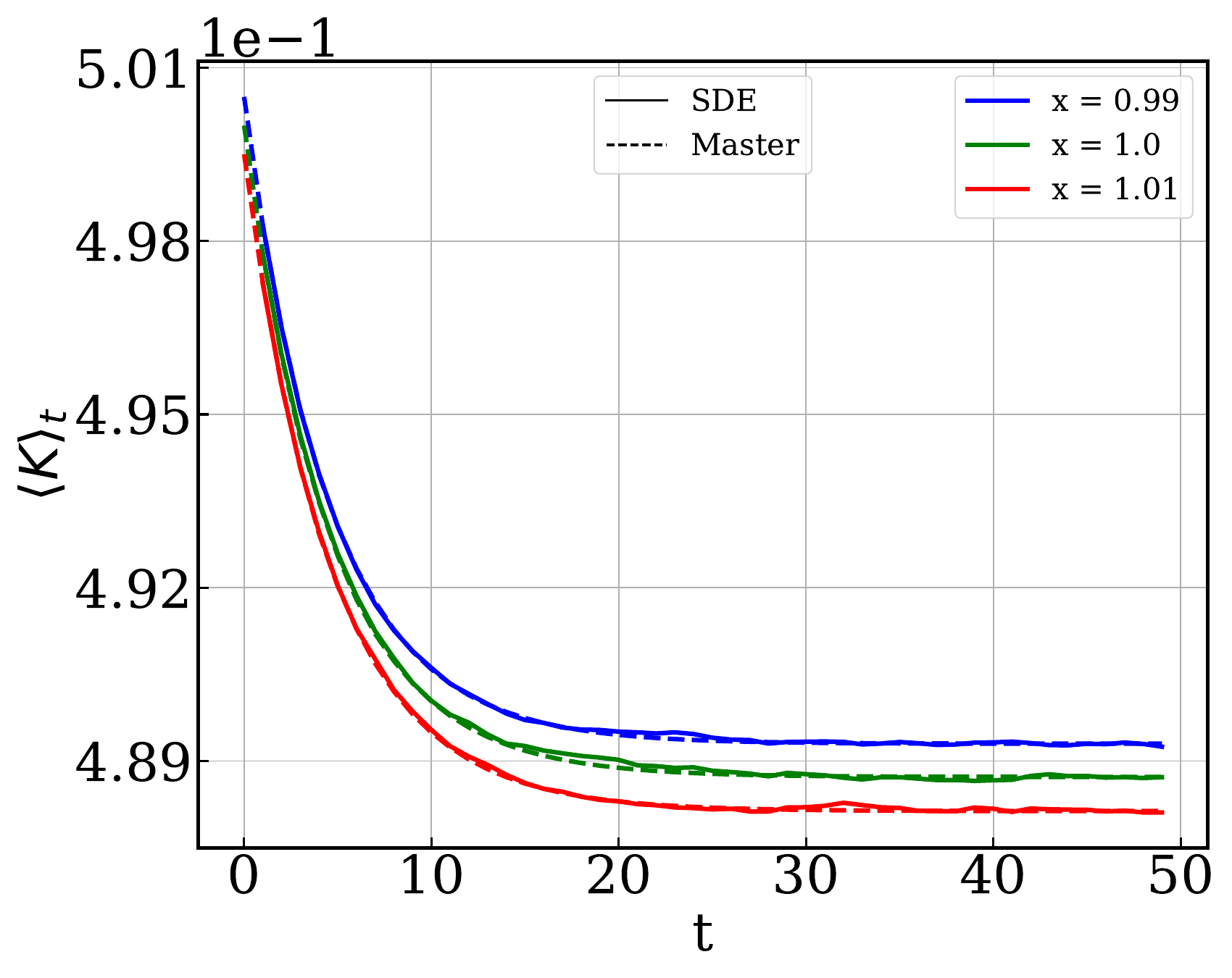}}
\caption{Steady state $\Delta^*$ as a function of confirmation bias $x$, for different values of (a) $\beta$ and (b) $p$ values. (c) shows the average action switching probability $\langle K \rangle_t$ as a function of $t$, for different values of confirmation bias. } 
\label{fig:delta_beta_p}
\end{figure}

We solve for the steady state the system of equations in Eq. \ref{eq:moments_evolution_final} and obtain a quadratic equation in $\Delta^*$. Upon taking the derivative of $\Delta^*$ with respect to the normalized confirmation bias\footnote{As proposed in \cite{palminteri2017}} $C = \frac{\alpha_+^c - \alpha_-^c - \alpha_+^u + \alpha_-^u}{\alpha_+^c + \alpha_-^c + \alpha_+^u + \alpha_-^u}$, we obtain that $\frac{\partial \Delta^*}{\partial C} > 0$ as long as $p(1-p) \neq 0$. This confirms our intuitions \cite{palminteri2017,Lefebvre2022} that confirmation bias leads to larger differences between the $Q$ values of the arms and therefore decreases the action switching probabilities. We are further able to note that $\frac{\partial^2 \Delta^*}{\partial \beta\partial C} > 0$, therefore implying that increasing the inverse temperature amplifies the effect of confirmation bias on the action switching probabilities. These trends can be seen in Fig. \ref{fig:delta_beta_p} (a,b) where we see $\Delta^*$ as a function of $x$\footnote{We use $x$ to parameterize a curve where $\alpha_+^c = \alpha_-^u = 0.1x$ and $\alpha_+^u = \alpha_-^c = 0.2-0.1x$, making $C = 2(x-1)$. }. In the figure $x=1$ represents unbiased learning rates and we have confirmation bias for $x>1$. We also note that for $x=1$, in Fig. \ref{fig:delta_beta_p} (a), the curves for all values of $\beta$ intersect at the same point. This is because for symmetric learning rates, the learning dynamics become decoupled from the action selection. This can also be seen from the analytical expressions in Eqs. \ref{eq:moments_evolution_final} by setting $\alpha^v_\pm = \alpha$. In particular, as Bayesian inference is unbiased we see that in Eqs. \ref{eq:moments_evolution_Bayes}, there is no dependence on action selection $i.e. 
 \beta$.

Nonetheless, we can conclude that introduction of a confirmation bias increases $\Delta^*$ thereby leading to lower $\langle K \rangle_t$ as can be see in Fig. \ref{fig:delta_beta_p} (c). 

\subsection{Action switching under time-varying learning rates}\label{sec:time_vary_argument}
Eqs. \ref{eq:moments_evolution_final} are greatly simplified for the case of Bayesian inference, as $\alpha_\pm^v = \alpha_t = \frac{1}{t+3}$. Making the appropriate substitutions, we obtain the following set of evolution laws.\begin{subequations}\label{eq:moments_evolution_Bayes}
  \begin{equation}
    \label{eq-aBayes}
      \langle Q_1 \rangle_{t+1} =  (1-\alpha_t ) \langle Q_1 \rangle_t 
+ p{\alpha_t}  ,
  \end{equation}
  \begin{equation}
    \label{eq-bBayes}
    \langle Q_1Q_2\rangle_{t+1} = (1-\alpha_t)^2 \langle Q_1Q_2\rangle_{t} + 2p\alpha_t(1-\alpha_t)\langle Q_1\rangle_{t}  + p^2{\alpha^2_t},
  \end{equation}
    \begin{equation}\label{eq-cBayes}
        \langle Q^2_1\rangle_{t+1} =  (1-\alpha_t)^2 \langle Q^2_1\rangle_t + 2p\alpha_t(1-\alpha_t) \langle Q_1\rangle_t + p\alpha^2_t.
  \end{equation}
\end{subequations} These evolution laws imply that $\Delta_t$ evolves according to
\begin{equation}\label{eq:bayes_delta_evolve}
    \Delta_{t+1} = \underbrace{(1-\alpha_t)^2\Delta_t}_\text{\clap{epistemic drift~}} +  \overbrace{p(1-p)\alpha^2_t}^\text{\clap{noise~}}, 
\end{equation} where $\alpha_t = \frac{1}{t+3}$ as seen in Eq. \ref{eq:effective_learning_Bayes}. From Eq. \ref{eq:bayes_delta_evolve} is particularly insightful as one can identify two distinct components to the time evolution of $\Delta_t$: a drift term and a second noise driven term. We take note that the drift term has a coefficient $(1-\alpha_t)^2 \leq 1$, implying that the drift always tries to drive $\Delta_t$ to zero. This makes sense, as we are dealing with symmetric environments and Bayesian inference indeed converges to the truth asymptotically i.e. $\Delta_{\infty} = 0$. The noise driven term is proportional to the variance of the environment $p(1-p)$ which drives $\Delta_t$ to larger values (proportional to $\alpha^2_t$). This allows us to guess how Bayesian inference sustains low $\langle K\rangle_t$ for the time-scales observed in experiments - in the small time regime because we start from $\Delta_0 = 0$, the drift term is small compared to  the noise term. Therefore, random fluctuations of the environment drive the $Q$ values apart, decreasing the action switching rates. But at slightly larger times, the noise term becomes insignificant as it is quadratic in $\alpha_t$ and the drift term is linear in $\alpha_t$. This implies that $\Delta_t$ starts reducing, albeit at a very small rate, because for small $\alpha_t$, $(1-\alpha_t)^2$ is close to unity. In totality, Bayes inference maintains low action switching rates, by first driving the $\Delta_t$ to higher values by environmental fluctuations, and then by decreasing the learning rates such that the $Q$ values become "sluggish", thereby sustaining the low action switching rates.

\begin{figure}
\centering
\subfloat[]{\includegraphics[width = 1.7in]{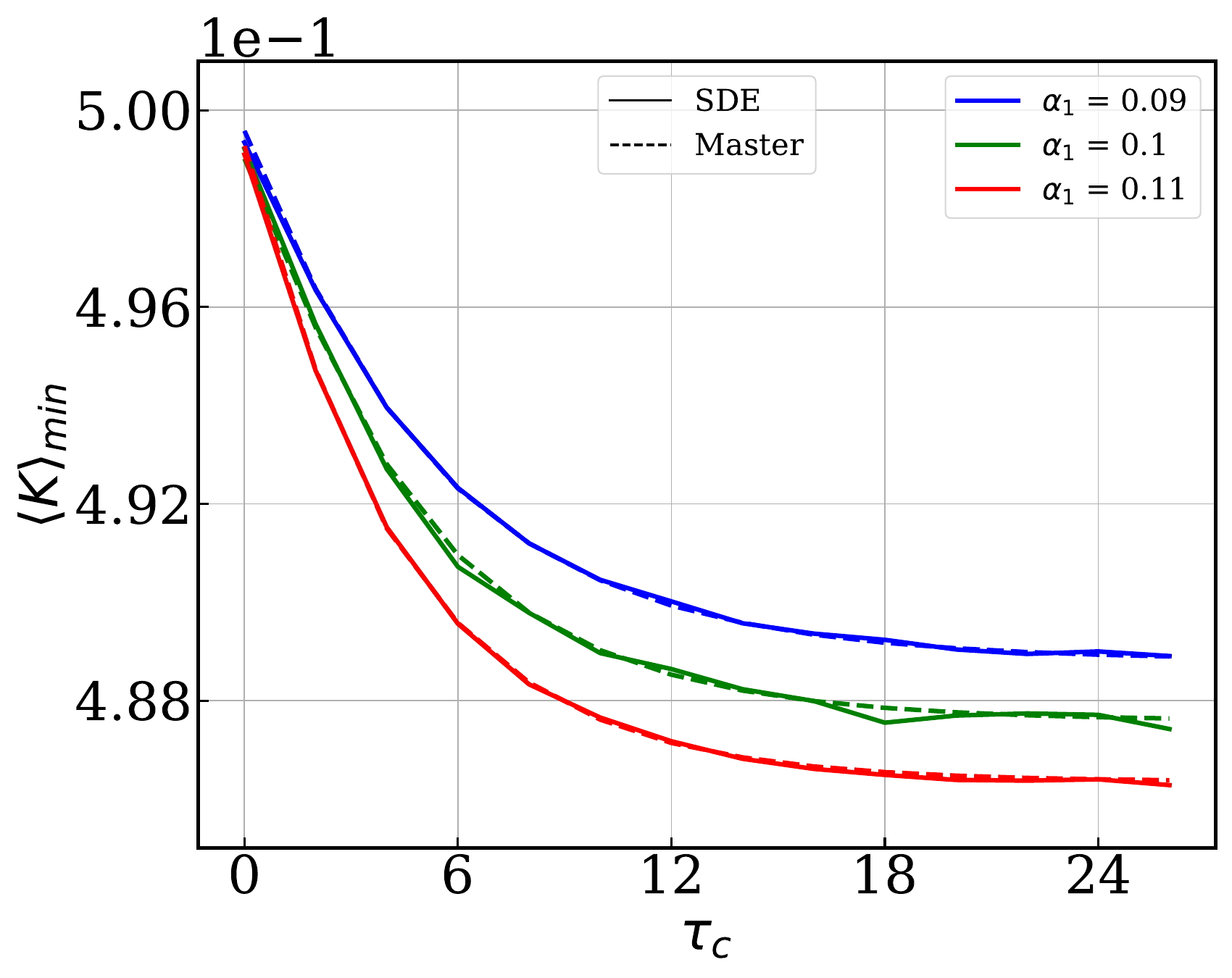}} 
\subfloat[]{\includegraphics[width = 1.7in]{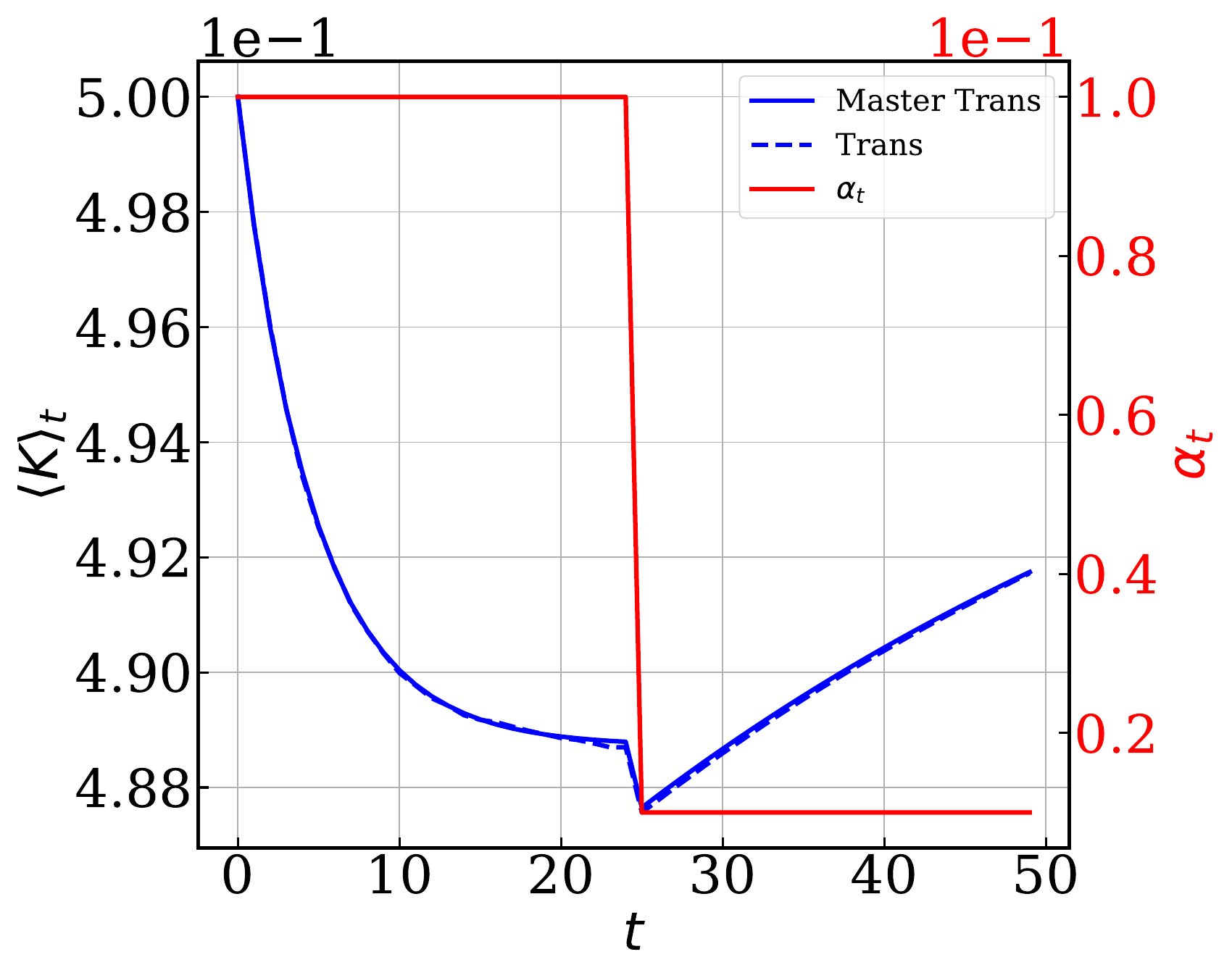}}
\caption{(a) The minimum (across time $t$) value of the average action switching rate $\langle K\rangle_{\text{min}}$ as a function of $\tau_c$ for different values of $\alpha_1$. (b) $\alpha_t$ and the corresponding action switching rates $\langle K\rangle_t$ as a function for time for $\tau_c = 25$ and $\alpha_1,\alpha_2 = 0.1,0.01$. } 
\label{fig:action_switch_time_vary}
\end{figure}

 We now make the above arguments quantitatively precise, by considering the time-scales in the system. First, we notice that for a stationary $\alpha_t = \alpha$, $\Delta$ relaxes to the steady state with a time-scale $\tau = \frac{1}{\alpha(2-\alpha)}$ and the steady state is given by $\Delta^* = \frac{p(1-p)\alpha}{ (2-\alpha)}$. As $\alpha \in [0,1]$, $\Delta^*$ is increasing in $\alpha$.  Additionally, $K$ is monotonically increasing in $\alpha$. This suggests that in order to obtain low $\langle K \rangle$ by modulating $\alpha_t$ alone, one first starts with a relatively large value for $\alpha$ and keep the value large for times comparable to $\tau$\footnote{As $\tau$ is decreasing in $\alpha$, this waiting period for large starting $\alpha_0$ is relatively small, as is the case for Bayesian inference where $\alpha_0 = \frac{1}{3}$.}. This will lead the system to have large $\Delta_t$. Once the $\Delta$ equilibriates, decrease $\alpha_t$ rapidly, thereby leading to small $\langle K \rangle_t$. Such a temporal protocol can sustain low action switching probabilities for relatively long times as $\tau$ is large for very small $\alpha$ and therefore $\Delta$ will decrease over a slow time-scale. For the purpose of demonstration we choose \[\alpha_t = \begin{cases}
    \alpha_1, & t<\tau_c \\ \alpha_2, & t\geq \tau_c,
\end{cases}\] and observe the impact on $\langle K \rangle_t$ in Fig. \ref{fig:action_switch_time_vary}. We are able to notice that $\langle K \rangle_t$ experiences a sudden drop at $\tau_c = 25$ and then slowly approaches (a higher) equilibrium corresponding to $\alpha_2$. In doing so, for an intermediate time the action switching rates are lower than the case with $\alpha_2 = \alpha_1$. The minimum $\langle K \rangle_t$ depends on the exact value of $\alpha_1$. More notably achieving the same $\langle K \rangle_{\text{min}}$ is possible for smaller values of $\tau_c$ as $\alpha_1$ increases, thereby confirming our intuitions arising from Eq. \ref{eq:moments_evolution_Bayes}.

With the above arguments in place, it is relatively straightforward to see that we have identified two distinct routes to decreasing the action switching rates in bandit tasks. First, by introducing a bias in the learning rates. Second, by decreasing the learning rates with time, while being consistent with the arguments in Sec. \ref{sec:time_vary_argument}. In conclusion we have answered the question we set out for at the beginning of this section.

\section{Consequences on human behavior}
We have so far explored that both confirmation bias and Bayesian inference can lead to similar behavioral signatures.  The confounding effects of Bayesian inference and confirmation bias leads us to a further question - how do we distinguish bias from optimal inference or a more general temporal profile of the learning rates? Before we even begin to attempt an answer, we first take a note of how well these models explain human behavior in stationary TABB tasks.

\subsection{Model comparison}
We proceed by comparing model fits to data from \cite{palminteri2017}. When performing fits to data we consider, a Bayes-greedy model (Eq. \ref{eq:Bayes-greedy}, referred as "Bayes"), unbiased Q-learning model (Eq. \ref{eq:Q_learning} with $\alpha_\pm = \alpha$, referred as "Const.") and the full Q-learning model (Eq. \ref{eq:Q_learning}, referred as "Full") and also a "confirmation model agent" (Eq. \ref{eq:Q_learning} with $\alpha_+^c = \alpha_-^u$ and $\alpha_-^c = \alpha_+^u$, referred as "Conf."). We keep the fitting procedure to be similar to that of \cite{palminteri2017}- i.e. we minimize the negative log-likelihood of the models for each participant's data to find the best fit parameters. As the models vary in their complexities (degrees of freedom (represented as "df" in the following tables)), we obtain for each participating subject, the Bayesian information criterion (BIC) scores corresponding to each of the best-fit models. Table 1 shows the mean and standard deviations of the best fit negative log-likelihoods (NLL) and the BIC scores for each of the model for all the subjects.
\begin{table}[h!]
\centering
\begin{tabular}{|c|cccc|} 
\hline
 & \multicolumn{4}{c|}{\textbf{TABB with counterfactual information}} \\ \hline
 & \textbf{Bayes (1df)} & \textbf{Full (5df)} & \textbf{Const. (2df)} & \textbf{Conf. (3df)} \\ \hline
\textbf{NLL} & $9.48 \pm 6.21$ & $6.14 \pm 5.23$ & $9.12 \pm 6.29$ & $6.89 \pm 5.47$ \\ 
\textbf{BIC} & $22.13 \pm 12.43$ & $28.17 \pm 10.45$ & $24.59 \pm 12.57$ & $23.31 \pm 10.93$ \\ 
\hline
\end{tabular}
\caption{NLL and BIC values}
\end{table}
\begin{figure}
\centering
\subfloat[]{\includegraphics[width = 1.7in]{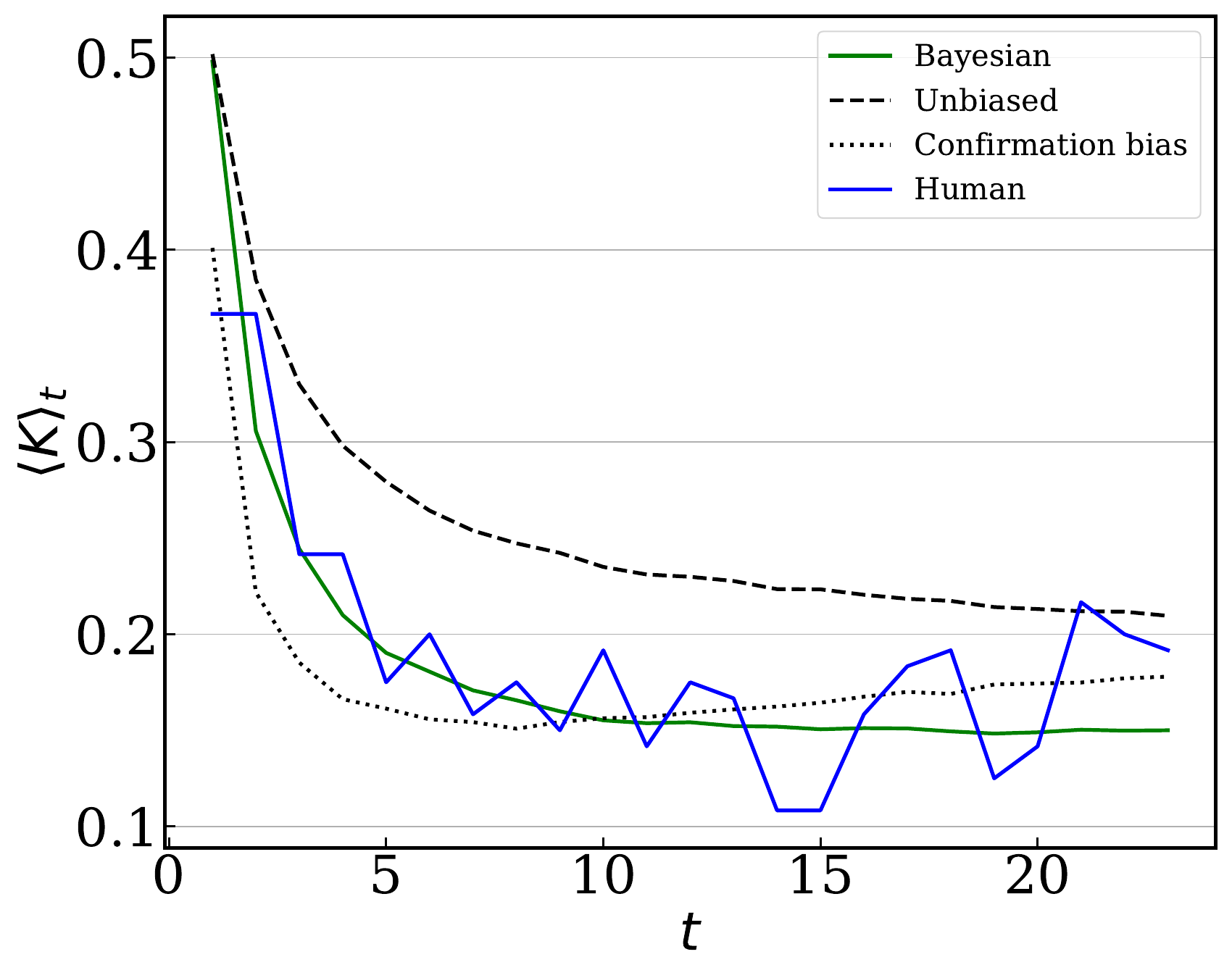}} 
\subfloat[]{\includegraphics[width = 1.7in]{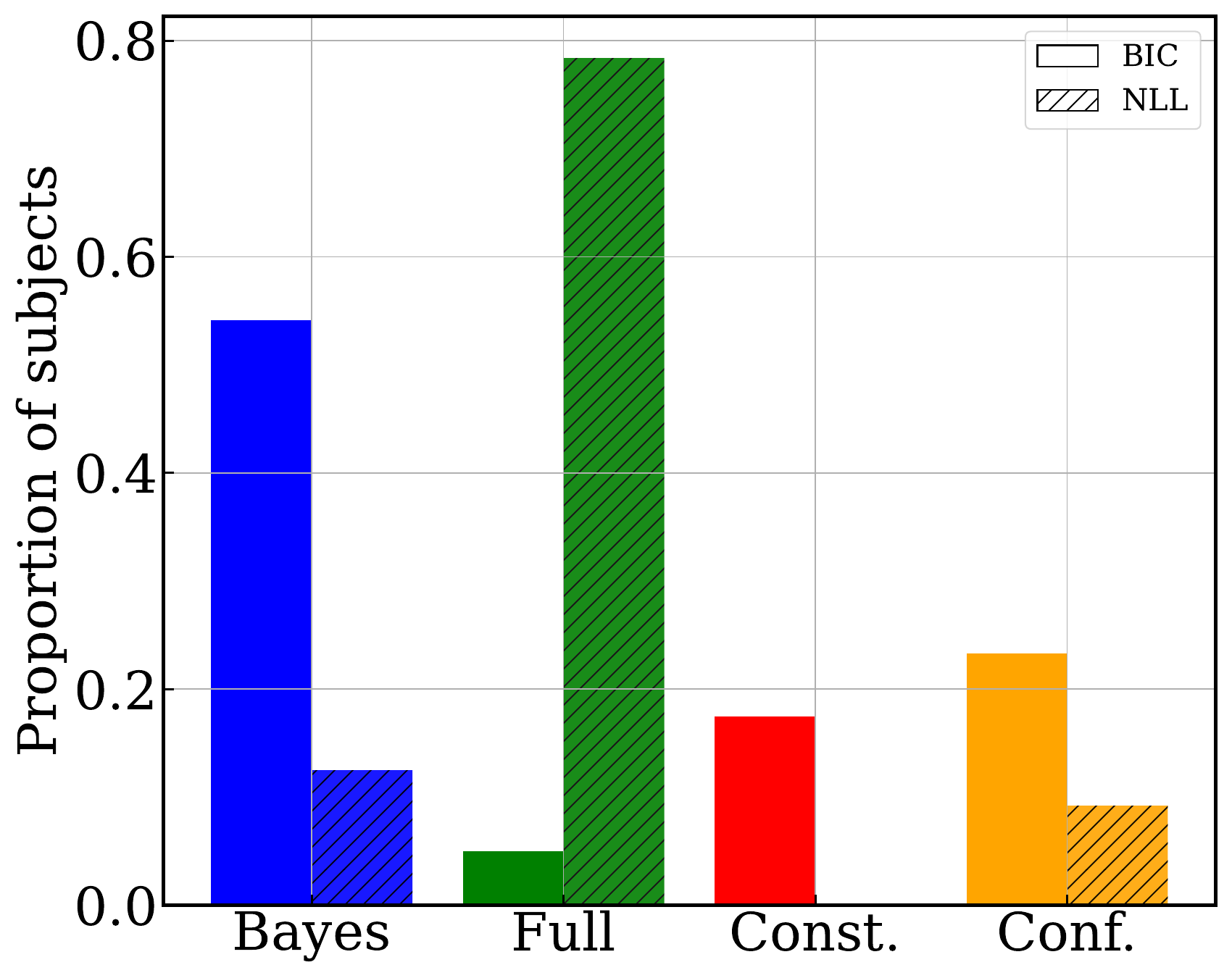}}\\
\subfloat[]{\includegraphics[width = 1.7in]{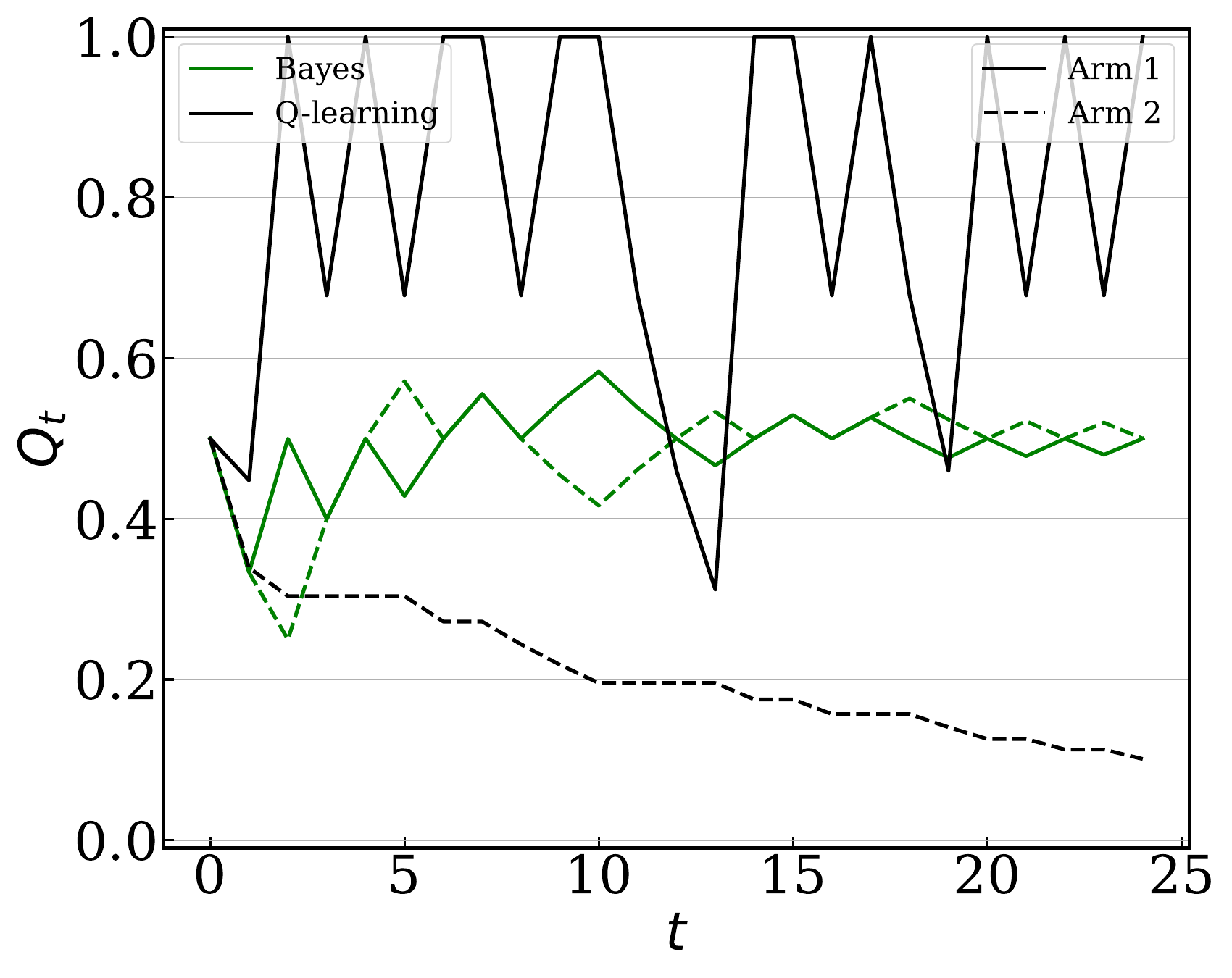}}
\subfloat[]{\includegraphics[width = 1.7in]{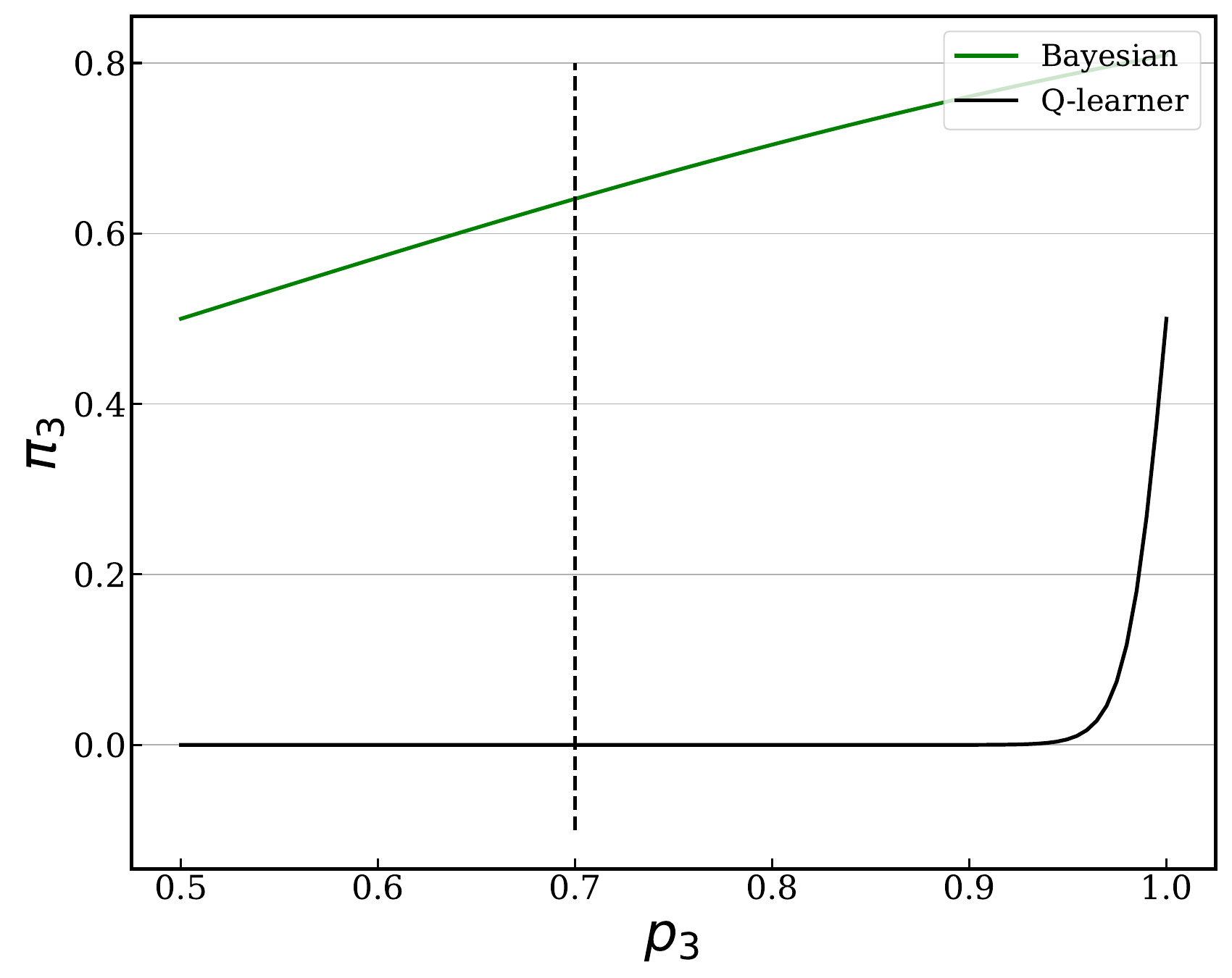}}
\caption{(a) Ensemble average action switching rates for human data (blue), best fit Bayesian agents (green), best fit $Q$-learning agents (black) for both unbiased (dashed) and biased (dotted) learning rates. (b) Distribution of best describing models for all human subjects by BIC (solid) scores and NLL (striped) scores. (c) Best fit $Q$-value trajectories for an examplar human subject corresponding to Bayesian (green) and $Q$-learning model (black). (d) Probability of choosing the newly introduced arm as a function of it's reward rate, for the best fit Bayesian (green) and $Q$-learning model (black).  } 
\label{fig:best_fit_models}
\end{figure}
In Fig. \ref{fig:best_fit_models} (a) we present the average action switching rates for humans and the corresponding best fit Bayesian and $Q$-learning agents (both biased and unbiased). It can be seen that the best fit $Q$-learners with unbiased and constant learning rates are  unable to account for the low action switching rates observed in human data. However, both Bayesian and confirmation biased $Q$-learners are able to fit relatively well to human behavior.

Additionally, we report the proportion of best describing models in the subject population in Fig. \ref{fig:best_fit_models} (b). We say that a subject is best described by a model $i$ if it has the lowest BIC of all the other models. As can be seen from Fig. \ref{fig:best_fit_models} (b), most subjects are best described by the Bayesian model, and quite expectedly by the "full" model if we ignore issues of model complexity.  As aforementioned, we have followed suit with \cite{palminteri2017} in so far as penalising model complexity is concerned, but we acknowledge that penalising complexity is not straightforward and therefore we do not take the above results to mean - "humans are Bayesian". However, these results suggest that if we account for temporally decreasing learning rates, we are likely to observe lower levels of confirmation bias as compared to when we do not account for it. In particular, behavior may be better described via hybrid models which have a combination of bias and temporally decreasing learning rates. Therefore, further investigation would be needed to disentangle Bayesian and biased models from human behavior in TABB tasks. 

\subsection{Potential empirical investigations}
As we explored in Sec. \ref{sec:time_vary_argument}, while both Bayesian and (confirmation) biased agents exhibit decreasing action switching rate $\langle K \rangle$, they do so via distinct mechanisms. Confirmation bias achieves low $\langle K \rangle$ by driving the system to larger $\Delta$ and Bayes inference achieves it by lowering $|\alpha_t|$. This suggests that on average the two mechanisms will lead to different steady state $Q$ values. In Fig. \ref{fig:best_fit_models}(c) we compare the $Q$ values obtained from the best fit Bayesian model and the best fit $Q$-learning model (at terminal time $t=24$) for an example participant in a symmetric environment (i.e. $p = 0.5$). We see that our two guesses for human behavior imply very distinct $Q$-values. In particular we see that the Bayesian model predicts that the $Q$-values converge to the true reward rates $p$, while the $Q$-learning model drives the $Q$-values further apart.

In order to empirically distinguish between the Bayesian inference and biased $Q$-learning, one may, at the end of the task, introduce a new arm with a reward probability $p_3$ which lies in between the Bayesian and $Q$-learning best fit $Q$ values for a given arm (say arm 1). The subject would have to be separately trained on the new arm and then asked to choose between the new arm and a given arm (arm 1), whether or not the player chooses the new arm might help distinguish between Bayesian and $Q$-learning agents. In Fig. \ref{fig:best_fit_models} (d) we present the supposed probability of choosing the new arm ($\pi_3$) as a function of it's reward rate $p_3$ for both the Bayesian and $Q$-learning model.

\section{Discussion}
This work present a rigorous theoretical comparison between $Q$-learning and Bayesian inference in TABB task with counterfactual information. We utilize our theoretical analysis to critique recent experimental studies on human behavior in TABB tasks. In particular we demonstrate that certain temporal profiles (including that of Bayes inference) of unbiased learning rates may show the behavioral signatures similar to that of a biased learning algorithm. This analysis also sheds light on why recent studies \citep{Lefebvre2022,Bergerot2024} observe that confirmation bias leads to better accumulated rewards in comparison to unbiased $Q$-learning, where they assume the constancy of learning rates.

While doing so, we've also demonstrated the usefulness of Master equation type formulations, commonly used in statistical physics, to study RL algorithms. An analysis similar to our steady state analysis, has been done already \citep{Lefebvre2022} without using the master equations. However their approach only works for epsilon-greedy algorithm with temporally constant learning rates, while our methodology is applicable to more general RL algorithms. In addition, the presented approach allows us to study the temporal dynamics of these learning systems, which cannot be obtained from a steady state analysis. 

In conclusion, we demonstrate that making claims about confirmation bias by fitting a $Q$-learning model as in Eq. \ref{eq:Q_learning} to behavior can be problematic. As this methodology has been employed by lot of studies in cognitive science literature, we believe that our analysis will allow cognitive scientists to make more nuanced and robust claims about confirmation bias when inferred from decision making tasks.

\bibliography{biblio}
\end{document}